\documentclass[conference]{IEEEtran}
\IEEEoverridecommandlockouts
\usepackage{cite}
\usepackage{amsmath,amssymb,amsfonts}
\usepackage{algorithmic}
\usepackage{graphicx}
\usepackage{textcomp}
\usepackage{algorithm}   
\usepackage{listings}
\usepackage{xcolor}
 \usepackage{url}
\usepackage[hidelinks]{hyperref}
\def\BibTeX{{\rm B\kern-.05em{\sc i\kern-.025em b}\kern-.08em
    T\kern-.1667em\lower.7ex\hbox{E}\kern-.125emX}}

\begin{document}

\title{Neuromorphic Control for 3D Navigation in Minecraft Using Genetic Algorithms
\\

}

\author{
\IEEEauthorblockN{Eric Zipor}
\IEEEauthorblockA{\textit{Department of Electrical and Computer Engineering} \\
George Mason University, Fairfax, VA, USA \\
ezipor2@gmu.edu}
}

\maketitle

\begin{abstract}
The popular 2009 voxel based videogame, Minecraft, contains several distinct  disciplines. One of which is ``parkour,'' gameplay that focuses on traversing a world's environment with maximum efficiency. The Minecraft online community has turned the game's physics engine into dynamic puzzles, requiring players to masterfully manipulate motion mechanics through frame precise timing of keystrokes. Actions such as sprinting, sneaking, and mouse direction are all combined to clear specific difficult jumps. Through this project, we design a genetic algorithm to generate weights for a neural network to autonomously evaluate inputs for block distances, terrain, and obstacles to determine the most optimal pathing.

\end{abstract}


\section{Introduction}
This project's potential extends beyond the niche video-game scene. ``Simulation-to-reality'' (Sim2Real) is a growing trend in robotics, and although Minecraft is not a realistic movement engine, the fundamental concepts of trajectory, and spatial awareness are still present. By training a ``brain'' in a low stake virtual environment, we can develop upon robust navigation strategies that could eventually apply to real world unmanned systems such as drones and cars without the cost of material.

Gamers infamously like to break the boundaries of what they once thought was impossible. Speedrunning is a sub-category that has captivated many audiences with pushing the limits on videogames to look for shortcuts and glitches that may benefit their world record run. Our approach however will not require the grinding and monotonous hours poured with the skill of a gamer. Instead, we are more akin to Tool-Assisted Speedrunning (TAS)  and are interested in seeing the theoretical limits of perfection when the human factor is removed.

For this project we will be using the Mineflayer API, a bot framework that enables automation and intelligence into the game. It's simple Python bridge support (node.js module) allows bots to collect information and respond to their environment based on per in-game (IG) tick responses [6]. This sensor data will feed into a genetic algorithm, which will iteratively evolve the bot's neural weights over thousands of generations to ``breed'' the best values for evaluating and clearing obstacles. Our goal is to take this one step further and apply an agent to a never before seen course and experiment it's adaptiveness.

To our knowledge, this has never been done before for Minecraft parkour. Pathing mods exist such as Baritone, however, they are deterministic A*-cartesian-style navigation, solving ``get from A to B'' with primitive traveling, not the desired optimization of momentum question we were asking [7]. We are also deliberately doing this in Mojang's Minecraft: Java Edition rather than forking to a clone. Most prior work in evolutionary algorithims ports the problem to a simplified pygame or Unity environment so the physics can be sped up. We choose to train inside the real game engine, on real server ticks, against the actual IG physics. This choice later cost wall-clock training speed but gave a genuine Sim2Real experience.

\begin{figure}[htbp]
\centerline{\includegraphics[width=0.9\linewidth]{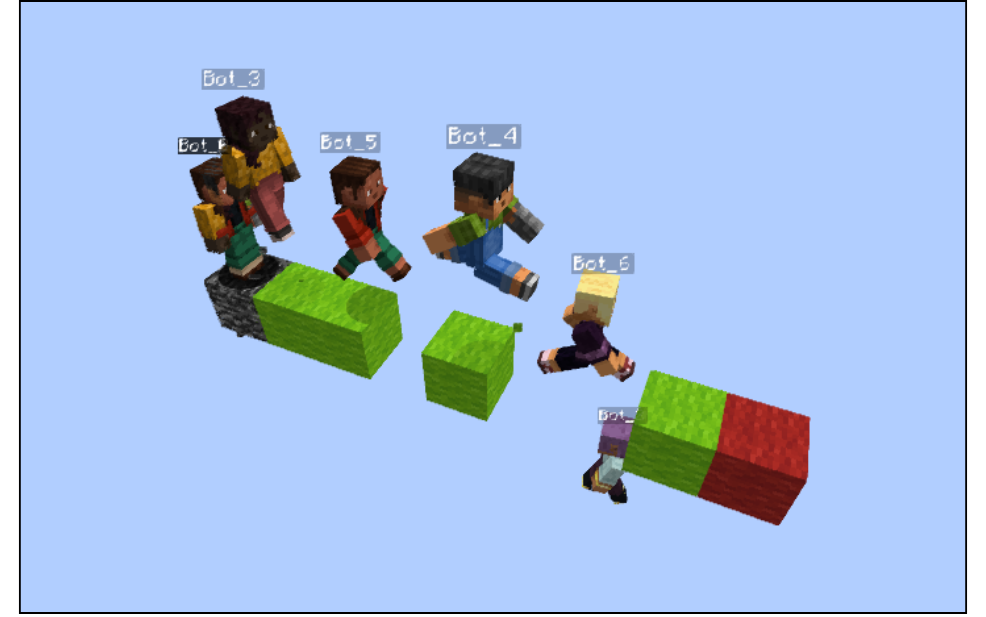}}
\caption{Parallel generation of bots navigating a randomized string of blocks and gaps on a local server. Spawning on bedrock (left), bots jump across or fall into the void, with the end goal of landing on the red wool (right).}
\label{fig1}
\end{figure}

\section{Setup}
To make the environment reproducible across runs we host a Vanilla Minecraft server, version 1.19.4, with no gameplay-altering mods. The server runs on a single laptop on a local area network, while bot clients connect from a separate desktop machine over the LAN. The world spawn is fixed to a coordinate and has a several in-game rules changes that would otherwise interfere with training (server.properties). Fall damage is turned off so a failed jump does not require a respawn animation, `doImmediateRespawn` is enabled, `randomTickSpeed` is set to zero so world elements do not change underneath the bots, and `sendCommandFeedback` is disabled to keep the server log clean. All bots are added to a single team with `collisionRule` toggled never and `friendlyFire` false so they do not push and are invisible to each other.

Each bot receives the same set of sensor readings every server tick and emits the same four control outputs, which Mineflayer translates into keypresses (forward, jump, strafe-left, strafe-right) and a mouse direction (yaw delta). From the server's perspective each bot is indistinguishable from a human player.

The last piece of infrastructure in the world is a chain of command blocks in the game driven by entity markers tagged `block\_pos` and `rng\_source` that randomize the layout of the obstacles between generations. The description of that  critical enhancement is referenced later in Section V, the driving mechanism for our random curriculum learning strategy.

\section{Neural Network Design}
\subsection{Input Layer}
Our first input layer prototype encoded the bot's surroundings as a voxel scan: a fixed 3D window of blocks around the bot, flattened into a one-hot vector. This gave the policy ``perfect'' information, however, the input layer would balloon past over a thousand neurons and the per-tick cost of querying every block from the Mineflayer world view dominated the digestion loop. The genetic algorithm needed to evaluate hundreds of policies per generation, so a single bot taking even milliseconds longer per tick could compound cost and error across a training run.

To solve this, we replaced the voxel grid search with simple raycasting. Each ray is cast from the bot's eye position out to a maximum distance of 10 blocks and returns a normalized hit value 1 - (hit\_distance / max\_distance), or zero if the ray misses completely.

\begin{figure}[htbp]
\centerline{\includegraphics[width=0.8\linewidth]{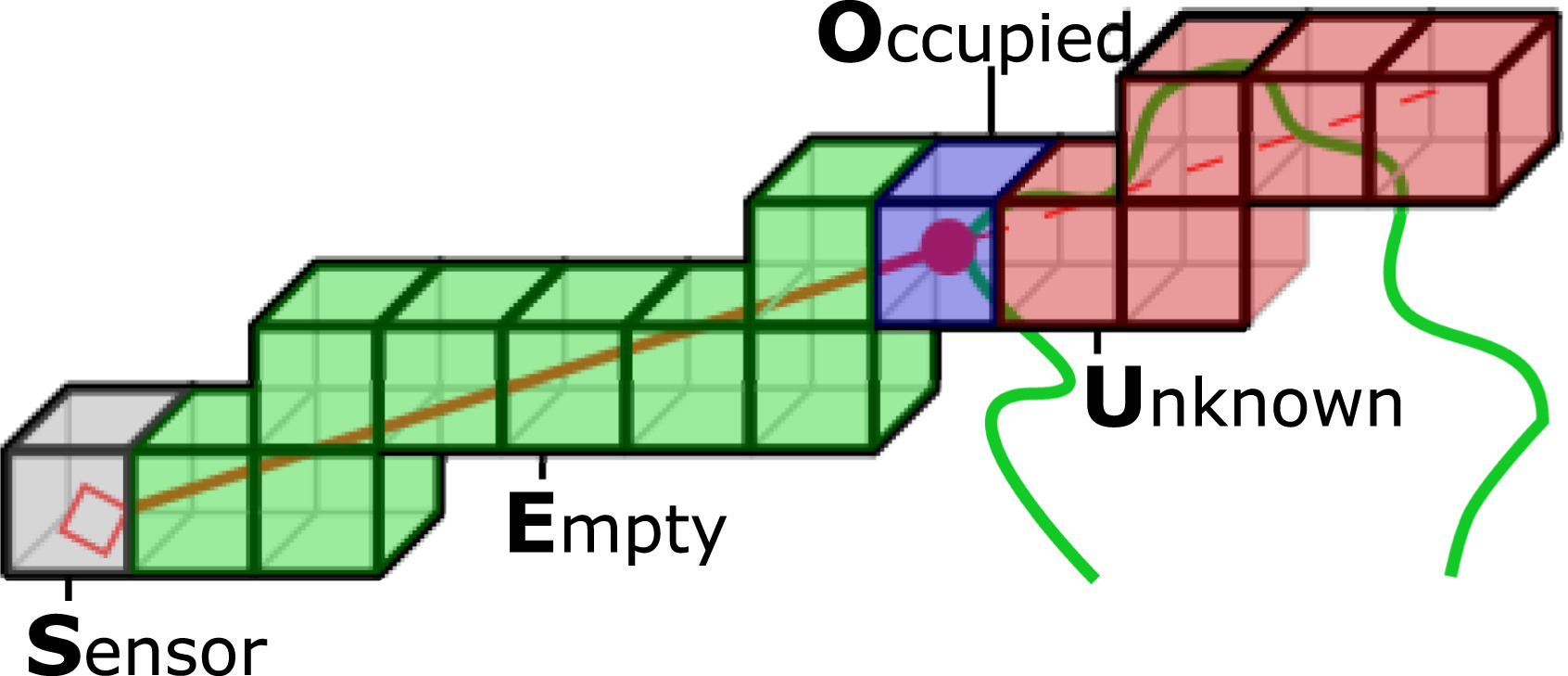}}
\caption{Example of occlusion detection via voxel status using ray casting from the sensor (grey cube) to the measured point (red dot) [3].}
\label{fig2}
\end{figure}

Raycasting has two distinct advantages. First, the input vector shrinks from thousands of neurons to just 19, which is a small tractable search space for the genetic algorithm. Second, raycasting more closely mirrors what a human parkour player actually has access to which is a line-of-sight (LOS) read on the geometry directly in front of them, not an oracle map of every block existing. The bot, like the human player, cannot foresee obstacles outside its forward hemisphere. The agent has to commit to a trajectory and discover the consequences.

The final input array consists of 33 neurons, 19 rays panning 180 degrees forward, seeking ground, ledge, and ceiling detection, the goal's offset as an (X,Y,Z) vector from the bot's respective frame; the bot’s velocity, if on-ground flag, the distance to the goal, and finally, an internal clock to encourage efficiency.

\subsection{Hidden Layer}

During development, the model architecture remained a single-hidden-layer multilayer perceptron (MLP) with 32 neurons. Increasing the number of hidden neurons could help mitigate underfitting by providing additional learning capacity, however, excessive neurons may lead to overfitting, expanding the genetic algorithm search space, and increase per-tick Mineflayer API call cost. Ultimately, 32 hidden neurons was good enough to showcase expressivity during experiments.

\subsection{Output Layer}
The output layer only has four necessary neurons! Three of the four outputs are binary, on or off actions: strafe left, strafe right, and jump (on a keyboard setup: buttons ``A'',``D'', and ``Space bar''). The fourth output is a continuous mouse yaw delta. We mix the activation functions on the output layer. Sigmoid squashes a real number into the interval (0, 1). For the three binary actions we threshold the sigmoid at 0.5 and treat the result as ``key pressed'' or ``key not pressed.'' The bot simply is either holding the jump key this tick or it isn't. The tanh activation function squashes a real number to (-1, +1). For the mouse angle we want a signed continuous value -1 for a hard left turn, +1 for a hard right, or 0 for looking ``straight.'' Sigmoid forces the bot to choose, tanh can commit to a magnitude and a direction in the same number. 

We locked sprinting forward for every tick the bot is alive. This is a deliberate simplification of the action space as parkour rewards constant commitment towards the destination, and giving the network the option to stand still would produce coward policies.

\section{Evolution}

\subsection{Genetic Algorithm}

The population is a list of N neural networks with identical topology and randomly (Glorot) initialized weights. Every generation, a network is loaded into a bot, the bot runs the obstacle course for a fixed timeout, and a fitness score is recorded. Selection is tournament-based with a tournament size that scales with the population. See equation 1 below.

\begin{equation}
T(N) = \mathrm{round}\left(2 + \sqrt{N}/2\right)
\end{equation}

The top elite\_count networks survive into the next generation unchanged and the remaining slots are filled by breeding by per-gene Gaussian mutuation. If the best fitness fails to improve by more than 50 points for three consecutive generations, the mutation rate is doubled to escape the plateau. Once improvement resumes they snap back to their initial values.

\subsection{Fitness Function}

The fitness function is designed so that a bot does not need to fully reach the goal in order to receive reward. Instead, it assigns based on progression, such as exploring new areas and staying alive while continuing to improve upon its position. It encourages efficient motion and discourages stagnant or repetitive behavior like circling or jumping without purpose. As the bot gets closer to the coordinate of the goal, the reward increases more sharply. A substantial amount of points is given when the bot finally touches the goal, along with an additional bonus reward when there is a new record for fastest completion.
The system reduces score based on unproductive behavior, such as dying early, or getting stuck for the entire duration. This ensures that ineffective strategies do not dominate the learning process, and the evolutionary algorithm is pushed consistently toward behaviors that improve the performance over time.

\subsection{Parallel vs. Serial Evolution}
In principle every member of the population could be evaluated in parallel: spawn N bots, run them all simultaneously, collect N fitness scores in one timeout window. This is what we initially built.
In practice the limitation of network jitter killed it. The laptop server host (16gb of ram) owned all world events including the obstacle course geometry, the tick response rate, and every bot's position. While a separate personal computer running the genetic algorithm calculated (matmul) bot movement commands over the router. 

With one solo bot connected the round-trip was steady; with mass bots issuing sensor and control packets every tick the server would processes a jump command on tick t+1 instead of initial t. A parkour jump that requires frame-perfect input simply failed. Worse, the failures were non-deterministic as the same network would clear a jump in one trial and misfire in the next, purely because of delayed packet arrival. The fitness signal became dominated by network noise rather than weight quality.

Mitigating this would have required collocating the control loop with the server (an ``edge'' deployment) and rearchitecting and modding the game-engine boundary [2]. To avoid this, we instead serialized pushing one bot at a time, each network evaluated for several times, their fitness averaged across trials. 

\begin{figure}[htbp]
\centerline{\includegraphics[width=0.9\linewidth]{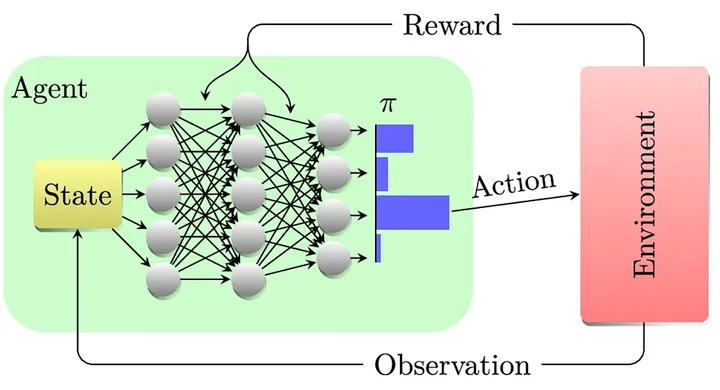}}
\caption{Example flow diagram of reinforcement learning methodology [5].}
\label{fig3}
\end{figure}

\section{Continual Domain Randomization}

\begin{figure}[htbp]
\centerline{\includegraphics[width=0.975\linewidth]{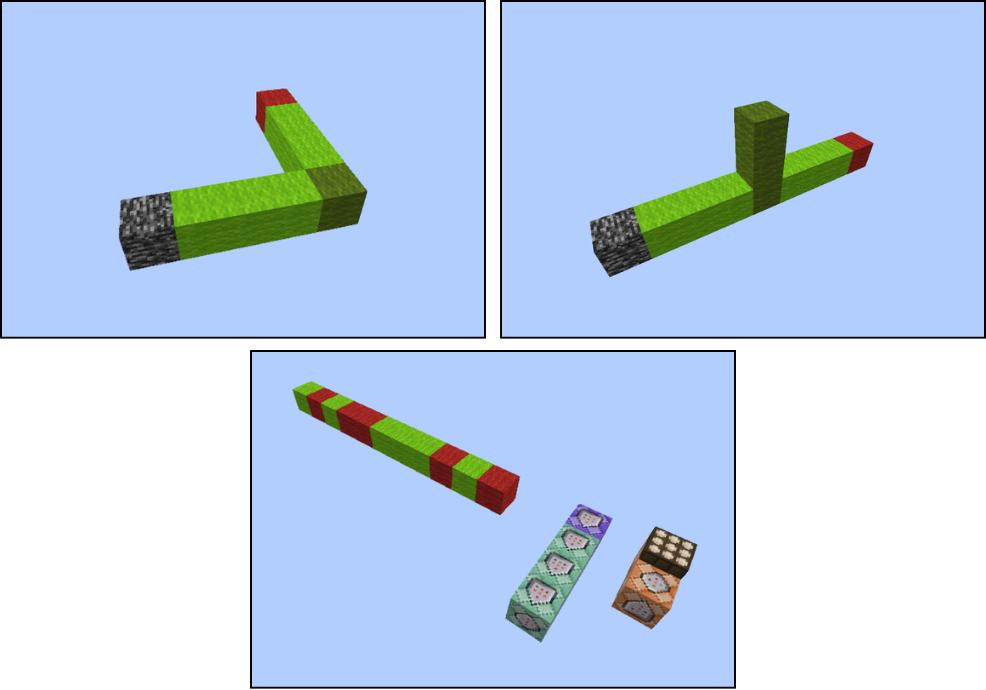}}
\caption{90 degree bend obstacle (upper left), jump-around-pillar, community recognized as a ``Single Neo'' (upper right), chained in-game command blocks summon a randomized balance beam after set amount of agent deaths, void gaps pictured here replaced with red wool (bottom middle).}
\label{fig4}
\end{figure}

After less than 100 generations on a fixed course, the networks would reliably clear the curated obstacle. We saved their best run weights in JSON format, indexed by which course they had mastered, a 90 degree left bend got its own file where the agent would cut diagonally, a jump-around-pillar (also known as a ``Single Neo'') agent apexes the turn balancing speed and precision. When it came to then trialing the same weights on a never before seen different layout the bot fall off immediately. Mirroring the course or adding a single extra block would result in poor performance. Our mistake quickly surfaced: how could the bot ever expect to turn right when it has only learned left?

What the network had actually learned was a motor sequence. When the agent's internal clock hit a certain tick on the server, it'd pivot 90 degrees counter-clockwise and jump. It had no understanding of parkour; it rather was memorizing a motor sequence, overfitting to one course. Being able to save off the weights per the specific course lets you only deploy to that exact track, it does not give you the overarching goal for a single network that could one-shot any obstacle course.

The fix is Continual Domain Randomization (CDR). This principle is rooted solving machine learning computer vision (object detection, pose estimation) and robotic control. The idea is to make training environments so variable that simulation-specific shortcuts are no longer reliable, forcing the network to learn the task specific relevant features [1]. By crafting a randomized curriculum, models learn to ignore the environment and focus on core task at hand.

Our implementation lives entirely inside Minecraft as a chain of command blocks. We summon a row of marker entities tagged block\_pos, one per cell of the obstacle course. Two more markers tagged rng\_source hold scoreboard values 0 and 1: one for ``place a block here'' and one for ``leave this gap empty.'' Each generation, after a configurable death threshold is met, a repeating command block randomly assigns each block\_pos marker one of the two rng\_source values, and chained command blocks then either place a lime-wool block or clear the cell to air. The course is regenerated from scratch every time the threshold trips, and no two generations see identical geometry.

To keep the search problem tractable we refined the scope to less complicated obstacles. Instead of full multi-turn parkour courses, the randomized obstacle is along the z-axis, and the bot only has to learn jumps of 1 to 4 blocks across arbitrary gap patterns (See Fig 1). Within that scope, CDR worked perfectly. After training under randomization the network generalized to balance beam style obstacles it had never seen, including gap patterns outside the training distribution.

\section{Future Work}

In the future, I'd like to incorporate hierarchical reinforcement learning. A single 32-neuron MLP is too demanding to handle every situation with one shared set of weights. Self-driving systems do not work this way, for example: Tesla cars do not use a single network for highway merging, reverse parking, and roundabouts; instead it dispatches to tailored networks [8]. To have a RL meta-controller that observes the bot's situation and routes to a specialist policy would address catastrophic forgetting: a specialist's weights are not at risk of being overwritten by a curriculum that has moved on to a different sub-task.

A second direction is curriculum learning under human guidance. The CDR randomization in Section V is uniform: every gap pattern is equally likely 5050. A more sophisticated training loop would let a human (or a meta-learner) dynamically scale the difficulty distribution, sampling more often from gap configurations the current population is failing on. The Minecraft setting has an unusual advantage here. Most AI researchers use static environments with pre-compiled physics simulations where dynamic curriculum changes are expensive; by using Minecraft we effectively have a real-time, 3D interactive sandbox where the curriculum can be edited live with command-block changes, no recompile and no engine restart. Whether dynamically scaling, human-guided curriculum results in faster convergence than uniform CDR is for open question.

Finally, the network-jitter problem from Section IV is itself a research direction rather than just a nuisance. The intersection of consumer-grade network latency and real-time reinforcement-learning action selection is under studied, most RL benchmarks assume zero-latency action delivery. My setup is accidental in when that assumption breaks. A follow-up could be characterizing how jitter distributions affect policy convergence, and whether a policy can be trained to ignore it, similar to CDR.

\section*{Acknowledgment}
A special thanks to Dr. Maryam Parsa for her guidance and expertise on Neuromorphic Computing.

\end{document}